\documentclass[a4paper]{article}

\usepackage[utf8]{inputenc}
\usepackage[T1]{fontenc}

\usepackage{gb4e}
\noautomath 

\usepackage[margin=1in]{geometry}
\usepackage{setspace}

\usepackage{authblk}

\usepackage{amsmath}
\usepackage{amsthm}
\usepackage{amssymb}
\usepackage{mathrsfs} 
\usepackage{tipa}     
\usepackage{booktabs} 
\usepackage{graphicx}
\usepackage{longtable}
\usepackage{array}
\usepackage{caption}
\usepackage{subscript} 

\usepackage[style=authoryear-comp, backend=biber, natbib=true]{biblatex}
\usepackage[colorlinks=true, citecolor=black, urlcolor=blue, linkcolor=blue]{hyperref}

\newcounter{tempcounter}
\newcommand{\argline}[3]{#1#2 #3}
\newcommand{\umrA}{\hspace{1em}}
\newcommand{\umrB}{\hspace{2em}}
\newcommand{\umrC}{\hspace{3em}}
\newcommand{\tabSC}[1]{\textsc{#1}}
\newcommand{\whitecmidrule}{\addlinespace[0.5em]}

\usepackage{lipsum}

\title{Dancing with Deer: A Constructional Perspective on MWEs in the Era of LLMs}

\author[1]{Claire Bonial}
\author[2]{Julia Bonn}
\author[3]{Harish Tayyar Madabushi}

\affil[1]{U.S. Army Research Lab}
\affil[2]{University of Colorado Boulder, U.S.A.}
\affil[3]{University of Bath, U.K.}
\date{} 

\begin{document}
\maketitle

\abstract{In this chapter, we argue for the benefits of understanding multiword expressions from the perspective of usage-based, construction grammar approaches. We begin with a historical overview of how construction grammar was developed in order to account for idiomatic expressions using the same grammatical machinery as the non-idiomatic structures of language. We cover a comprehensive description of constructions, which are pairings of meaning with form of any size (morpheme, word, phrase), as well as how constructional approaches treat the acquisition and generalization of constructions. 
We describe a successful case study leveraging constructional templates for representing multiword expressions in English PropBank. Because constructions can be at any level or unit of form, we then illustrate the benefit of a constructional representation of multi-meaningful morphosyntactic unit constructions in Arapaho, a highly polysynthetic and agglutinating language. We include a second case study leveraging constructional templates for representing these multi-morphemic expressions in Uniform Meaning Representation.  Finally,  we demonstrate the similarities and differences between a usage-based explanation of a speaker learning a novel multiword expression, such as \textit{``dancing with deer,''} and that of a large language model.  We present experiments showing that both models and speakers can generalize the meaning of novel multiword expressions based on a single exposure of usage. However, only speakers can reason over the combination of two such expressions, as this requires comparison of the novel forms to a speaker's lifetime of stored constructional exemplars, which are rich with cross-modal details.}

\section{Introduction} 
\label{sec:Intro}

At its inception, Construction Grammar (CxG) approaches have stood apart from other theoretical approaches to language because of their explanatory power when it comes to multiword expressions (MWEs). CxG is able to provide the machinery for elegantly modeling MWEs for two primary reasons. First: in CxG, the basic unit of language is the construction (Cxn), which is a pairing of meaning with a form, where the forms can vary across every level of language -- thus there are morphological forms, lexical forms, and phrasal forms. As a result, CxG provides a unified theory for accounting for meaning in multi-morphemic structures in precisely the same fashion as MWEs or transitive phrasal structures. Second: as a usage-based theory, CxG research demonstrates how frequency plays a role in facilitating extension but also preventing overgeneralization of flexible, partially productive Cxns. Thus, CxG can account for the seemingly idiosyncratic flexibility and productivity of MWEs, which has historically presented a challenge to the simplest Natural Language Processing (NLP) techniques of treating MWEs like ``words with spaces'' \citep{sag2002multiword}.  

To elaborate upon these strengths in a CxG account of MWEs, this chapter begins by detailing the tenets of CxG approaches, with special focus on the notion of a Cxn and the role of frequency in the storage and generalization of Cxns (Section~\ref{sec:MWEs-CxG}). We then turn to a description of how constructional templates have been used in modeling English MWEs in the PropBank project \citep{palmer2005proposition}, enabling theoretical advances in our understanding of the characteristics of MWEs, as well as computational advances in their detection and semantic interpretation (Section~\ref{sec:EnglishPropBank}). In this way, we show the benefit of having one unified treatment of both lexical and phrasal structures. Next, we turn to a description of multi-morphemic expressions in Arapaho, including efforts to develop Uniform Meaning Representation \citep{bonn2023umr} for Arapaho. This requires eschewing traditional notions of word-hood in favor of  Cxns, whose forms do not require any clear delineation between morphemes and words (as there are no morphological, input/output rules in CxG) (Section~\ref{sec:MWEs-MCLs}).  Thus, we show the benefit of one theoretical approach to account for meaning in multi-morphemic and lexical structures. 
Finally, we describe the parallels between a speaker's usage-based acquisition of grammar and the way that Large Language Models (LLMs) process text in order to acquire an impressive amount of linguistic competency (Section~\ref{sec:LLMs}). We also present two exploratory experiments on both novel and nascent English MWEs that demonstrate that although frequency effects in a speaker's grammar mimic priors in an LLM's pretraining, this does not result in fully functional linguistic competence (Section~\ref{sec:exp1},~\ref{sec:exp2}). This, too, is expected under CxG: whereas an LLM's ``usage'' is all through the processing of text, a person's usage is enriched by cross-modal association, resulting in vastly different meaning features paired up with a given form. We close with recommendations for future work in developing constructional resources that will help to bridge the gap between human linguistic competency and that of our NLP systems (Section~\ref{sec:conclusion}). 

\section{Accounting for MWEs in Construction Grammar}
\label{sec:MWEs-CxG}
In this section, we begin with the importance of idiomatic expressions, one subset of MWEs, in the historical development of CxG. We then turn to a detailed description of what Cxns are, as well as what features of Cxns are posited to be stored in the mental lexicon, or `constructicon'. We then turn to a discussion of the importance of frequency of use and the factors that both facilitate and constrain the extension or generalization of Cxn types to novel instantiations. Such extension leads to Cxns of varying level of abstractness, or `schematicity,' discussed next. We close this section with a description of constructional templates, facilitating the  representation of both fixed and flexible constructional slots. 

\subsection{CxG: a Theory for ``Peripheral'' Language Phenomena}
\label{subsec:peripheral}
Croft and Cruse write, ``It is not an exaggeration to say that Cxn grammar grew out of a concern to find a place for idiomatic expressions in the speaker's knowledge of a language'' \parencite*[205]{Croft_Cruse_2004}. A place for idiomatic expressions had not been satisfactorily found in the Generative linguistic tradition, e.g. \citet{chomsky1957syntactic}, in which idiomatic expressions had been relegated to the ``peripheral'' language phenomena, which seemingly neither cooperated with nor merited an elegant rule-based explanation of properties. 
We define idioms as a class of MWEs
that exhibit characteristics of lexical idiosyncrasy and syntactic creativity. 
Idioms are lexically idiosyncratic in that they are tied to certain words with
different degrees of fixedness. They exhibit syntactic creativity in that their 
behavior may not be accounted for by existing syntactic rules. These characteristics are especially problematic when viewed from a Generative perspective, which posits the autonomy of syntax and semantics -- semantic features are stored in the lexicon, while syntax dictates the combinatoric potential of lexical items. The Generative solution to the problem of idiomatic expressions was largely to posit that their non-conventional meaning requires speakers to store them as frozen, inflexible chunks in their mental lexicon. However, the majority of idioms are not frozen, showing some syntactic flexibility, such as the ability to be marked for the past tense or passivized: \textit{``The beans that Nathan spilled...''} Others are more fixed, resisting passivization: \textit{*``The bucket that Grandpa kicked...''}\footnote{Usages thought to be less felicitous to native speakers are preceded by `*'.} If idioms are to be stored in the lexicon, this would reasonably lead one to conclude that some syntactic information is also stored in the lexicon, with respect to what syntactic rules are permissible for which idioms. Thus, to accommodate idioms, the lexicon either had to include such syntactic information, or conversely, one could conclude that syntactic rules exhibit idiosyncratic lexical behavior -- sometimes being choosy as to which lexical items can fill certain slots, but according to semantic criteria (the idiomatic meaning of the words) as opposed to morphosyntactic criteria. This makes strict separation of syntax and semantics quite difficult to maintain. 

As a result of their special characteristics making them uncooperative with the Generative view of syntax and semantics, idiomatic expressions were relegated to the peripheral, minor aspects of language. But are such expressions peripheral? Although definitions of idiomaticity vary, making corpus counts inexact, studies suggest that MWEs, where multiple words function as a single meaning unit, make up anywhere from 30 to 50\% of an English speaker's vocabulary \citep{sag2002multiword}. What's more, idiomatic expressions interact in important ways with the rest of the grammar, demonstrating a continuum of compositionality in language from lexically fixed, `substantive' idioms on the one hand, and more lexically flexible, `schematic' expressions on the other hand. 

Fillmore, Kay, and O'Connor advocate for ``the view that the realm of idiomaticity in a language includes a great deal that is productive, highly structured, and worthy of serious grammatical investigation'' \parencite*[501]{fillmore1988regularity}. The authors go on to argue that an explanatory model of grammar should be able to account for ``syntactic configurations larger and more complex than those definable by means of single phrase structure rules,'' and delve into the expression \textsc{let alone} as a case study. The authors summarize their definition of idiomatic expressions as, ``something a language user could fail to know while knowing everything else in a language,'' where everything else is thought of as the grammar and vocabulary of the language, which does not enable computation of the special meaning of idiomatic expressions. We refer the reader to this article for a detailed catalogue of idiomatic expressions, organized according to syntactic and semantic properties, such as ``unfamiliar pieces unfamiliarly arranged,'' a category including the \textsc{Comparative-correlative}: \textit{``The more carefully you do your work, the easier it will get.''} Notice that this expression is syntactically idiosyncratic in that there is no clear constituent category for each of the juxtaposed \textit{``the...''} phrases, yet there are some constraints on the parallel nature of the phrases, preventing, for example \textit{*``The more carefully you do your work, the easier,''} whereas \textit{``The more, the merrier''} is felicitous. 

Similarly, in the case of \textsc{Let-alone}, the authors pinpoint the syntactic properties of the expression as well as the semantic properties, accounting for the flexibility of the expression as well as the requirement for certain kinds of parallel structures on each side of ``let alone,'' which itself is frozen, allowing for no morphological variation or intervening words: 
\begin{enumerate}
    \item \label{ex:letAlone} 
        \begin{enumerate}
            \item \label{ex:eatSquid-A} F<X A Y let alone B> \\
                `I doubt you could get Fred to eat squid, let alone Louise' 
            \item \label{ex:eatSquid-B} F<X A let alone B Y>\\
                `I doubt you could get Fred, let alone Louise, to eat squid'
        \end{enumerate}
\setcounter{tempcounter}{\value{enumi}}
\end{enumerate}

\noindent In this syntactic schemata, each variable is assigned certain semantic properties as well. A and B are the two compared elements, above \textit{``Fred''} and \textit{``Louise''}. X, representing \textit{``you could get,''} in combination with Y, representing \textit{``to eat squid''} above,  sets up a scale.  The two juxtaposed structures before and after \textit{``let alone''} are points on that scale. Finally, F above, corresponding with \textit{``I doubt''} carries a pragmatic meaning, indicating the speaker's strength of illocutionary force, where the strength of their belief in the first element (doubting that Fred would eat squid) is not as strong as their belief in the second element (doubting all the more that Louise would eat squid). 

In detailing the syntactic schemata for \textsc{let alone}, the authors demonstrate one kind of formalism for capturing the syntactic slots of a Cxn, their properties, as well as the semantics associated with those slots. Significantly, the authors lay out the ways in which the formalisms of Construction Grammar are able to provide the explanatory mechanisms for idiomatic expressions, such as \textsc{let alone}, as well as purely compositional expressions abiding by the more typical combinatoric rules of phrase structure. Indeed, Fillmore, Kay and O'Connor conclude ``the machinery needed for describing the so-called minor or peripheral constructions...will have to be powerful enough to be generalized to more familiar structures.'' \parencite*[534]{fillmore1988regularity}. Thus, CxG was developed precisely to have such machinery, enabling a uniform account of all parts of language. 

\subsection{Constructions: Form and Meaning}
In this section, we provide greater detail into what linguistic information characterizes Cxns. Constructions are form-meaning pairings; thus, in many ways, Cxns are quite similar to the Saussurean sign, which is a pairing of a concept and what is termed a `sound image,' where these items in turn can be thought of as the signified (e.g., a tree) and the signifier (e.g., \textit{``tree''}), respectively \citep{de2004course}. As a usage-based and cognitively plausible grammatical theory, CxG posits that our linguistic abilities are explained by domain-general cognitive processes, and that many aspects of language rely on our capacity for symbolic thinking, that is, our ability to store arbitrary pairings of form and meaning. 

Cxns reflect these pairings and are characterized by what is often called a `form pole' and a `meaning pole.' The form pole is characterized by any phonologically fixed elements as well as the relative locations of more flexible constructional slots (i.e. basic ordering or syntactic characteristics). The meaning pole in CxG is merely a shorthand for a complex concept, the details of which vary from speaker to speaker, given that each speaker will have different contextual associations with that concept and when it is referred to by that particular signifier. Thus, we define a Cxn as an arbitrary pairing of (phonological/syntactic) form and meaning that is stored in a speaker’s mental lexicon, or, as it is sometimes called in CxG, a speaker's `constructicon'. Hereafter, we will refer to the fixed elements as `substantive,' and the flexible elements as `schematic', following the terminology of \citet{hoffmann_2022}. 

Although there are a variety of different schools of CxG \citep{hoffmann2013oxford}, all agree that the Cxn, as defined above, is the basic unit of language. As such, Cxns are found at every level of language -- morphemes are Cxns, words are Cxns, phrases are Cxns. For example, \textit{``unfair''} is an instance of the \textsc{un-adjective} Cxn, where \textit{``un''} is the substantive element and the adjective is the schematic element. 
 The category of the English adjective is itself a distinct schematic Cxn with no fixed phonology, and \textit{``fair''} is a fully substantive, lexical Cxn. Similarly, \textit{``below the belt''} is both an English prepositional phrase Cxn (with a literal meaning when constituted by lower-level Cxns denoting a spatial relation and an item of clothing), but also a unique, idiomatic phrasal Cxn that also has a meaning pole akin to \textit{``unfair.''}

Whether an English speaker computes the meaning literally according to individual component Cxns or the idiomatic meaning of the phrasal Cxn depends upon context. The importance of context is, of course, not new or special to CxG. Indeed, if any linguistic generalization has been borne out by the trajectory of successful language modeling -- where we see increasing linguistic sophistication as increasing amounts of context are taken into account, up to the point of the web-scale pre-training data of LLMs -- it is that context is critical in interpreting meaning. However, what does set CxG apart, and usage-based theories more generally, is the emphasis on the importance of context that goes beyond linguistic and beyond the current context of a particular usage.  Usage-based theories emphasize that each token of linguistic experience updates and enriches a speaker's stored representation of a Cxn with a variety of cross-modal associations \citep{bybee2006usage}. For example, an English speaker stores details about where they were, what was going on, how comfortable s/he was physically, who else was there, and other details of the full sensory experience when they have heard \textit{``below the belt''} used idiomatically.

\subsection{Constructional Templates for MWEs}
Constructional templates, which represent both the fixed, substantive elements of Cxns as well as the flexible, schematic elements, provide a powerful tool for modeling partially productive MWEs. Such MWEs present a challenge for both modeling and detection because, as partially productive Cxns, their words are not fully fixed (as would be the case in an unproductive Cxn, such as \textit{kicked the bucket}); thus, one cannot simply use a ``words with spaces'' approach to entering them into a computational lexicon and then associating a special meaning with the detection of that fixed form \citep{sag2002multiword}. On the other hand, their formulation is not highly productive (as would be the case, arguably, for English referential Cxns, wherein the pattern of \textit{``a''} or \textit{``the''} + noun is compatible with a wide variety of nouns). Thus, one cannot posit a clear rule for accounting for all possible cases. However, as the productivity continues to allow for novel variants to enter the language, one can never hope to list all of the different variants in a static computational lexicon.  Furthermore, attempting to list all variants leads to overproliferation of the lexicon and ignores generalizations that can be made across all of the related expressions.  

Positing Cxns within the computational lexicon (or constructicon) models the fixed and flexible characteristics of a Cxn or MWE. For illustrative purposes, we include Table~\ref{tab:constructions} showing the name of a Cxn, it's level of schematicity, a prose meaning pole description, form pole description, as well as an example of each Cxn. The form pole notation in Table~\ref{tab:constructions} is drawn from \citet{hoffmann_2022}.

{
\footnotesize
\begin{longtable}{|p{1.9cm}|p{1.7cm}|p{1.9cm}|p{2.7cm}|p{2.1cm}|}
\hline
\textbf{Construction} & \textbf{Schematic / Substantive} & \textbf{Meaning Description} & \textbf{Form Description} & \textbf{Example}  \\ 
\hline
\endfirsthead
\caption[]{-- Continued} \\
\hline
\textbf{Construction} & \textbf{Schematic / Substantive} & \textbf{Meaning Description} & \textbf{Form Description} & \textbf{Example}  \\ 
\hline
\endhead
Let-Alone 
& Substantive - ``let + alone'' is frozen 
& Invokes a scale of an implicit property, where preceding slot A is on the lower end in comparison to following slot B 
& \textsc{Phonology:} \par /A$_{1}$ l{\textepsilon}t$_{2}$ {\textreve}'lo{\textupsilon}n$_{3}$ B$_{{4/5}}$ \par \textsc{Morphosyntax:} \par /XP$_{1}$ CONJ$_{2-3}$ XP$_{4/5}$
& {[}{[}A ceasefire,{]}$_{1}$ let$_{2}$ alone$_{3}$ {[}lasting peace{]}$_{4}${]}$_{5}$, will take long negotiation. \\
\hline
Comparative-Correlative 
& Partially Substantive - ``the'' + comparative ADJ/ADV\dots + ``the'' + comparative ADJ/ADV 
& Specifies a cause-and-effect and/or temporal relation between the first comparative slot, and the second comparative slot
& \textsc{Phonology:} \par
  /{[}\dh\textschwa$_{1}$ A$_{2}$ B$_{3}${]}$_{\text{C}}$$_{1}$ {[}\dh\textschwa$_{4}$ C$_{5}$ D$_{6}${]}$_{\text{C2/7}}$ \par
  \textsc{Morphosyntax:} \par
  {[}{[}\emph{the}$_{1}$ {[}Comparative Phrase{]}$_{2}$ REST-CLAUSE$_{3}${]}$_{\text{C1}}$ \par
  {[}\emph{the}$_{4}$ {[}Comparative Phrase{]}$_{5}$ REST-CLAUSE$_{6}${]}$_{\text{C2}}${]}$_{7}$
& But {[}the$_{1}$ longer$_{2}$ {[}this goes on{]}$_{3}$, the$_{4}$ worse$_{5}$ {[}his odds get{]}$_{6}${]}$_{7}$.\\
\hline
Conative 
& Mostly Schematic - ``at'' is fixed
& Indicates an attempt to transfer force from an agent to a patient 
& \textsc{Phonology:} \par /A$_{1}$ B$_{2}$ {[}{\ae}t $_{\text{C}}${]}$_{3/4}$
  \par \textsc{Morphosyntax:} \par {[}SBJ$_{1}$ {[}V$_{2}$ OBL:\emph{at}-PP$_{3}${]}$_{\text{VP}}${]}$_{4}$
& {[}{[}His belly{]}$_{1}$ pushed$_{2}$ {[}at the buttons of his white and blue Aloha shirt{]} $_{3}${]}$_{4}$. \\
\hline
Resultative
& Fully Schematic 
& Agent of the action denoted by the verb causes a patient to change / become a resulting state
& \textsc{Phonology:} \par
  /A$_{1}$ B$_{2}$ C$_{3}$ D$_{4/5}$ \par
  \textsc{Morphosyntax:} \par
  {[}SBJ$_{1}$ {[}V$_{2}$ OBJ$_{3}$ OBL$_{4}${]}$_{\text{VP}}${]}$_{5}$
& {[}{[}It{]}$_{1}$ jerks$_{2}$ you$_{3}$ awake$_{4}$ with the first sentence\dots{]}$_{5}$ \\
\hline
\caption{\label{tab:constructions} Constructions of varying schematicity. The Form pole description leverages constructional templates to distinguish phonologically fixed, substantive slots (denoted in IPA), and the morphosyntactic character of schematic slots (denoted with variables in the phonological description that are co-indexed with grammatical constituent types in the morphosyntactic layer). MWEs are either substantive, inlcuding Let-alone shown here, or partially substantive.}
\end{longtable}
}

\section{Constructional Rolesets for English MWEs in PropBank}
\label{sec:EnglishPropBank}
A constructional template, in which a form pole (as seen in Table~\ref{tab:constructions}) representation is paired up with shorthand prose denoting the meaning pole, is very similar to the ``rolesets'' for semi-productive MWEs taken on by the English PropBank semantic role labeling project \citep{palmer2005proposition}. Here, we turn to PropBank as a real-world use case benefited by leveraging constructional modeling of MWEs. 

The Proposition Bank (PropBank) uses syntactic parses as a scaffolding for the much more difficult problem of parsing meaning. The underlying idea was that English verbs exhibit patterns in the way they structure their participants both syntactically and semantically, and so by tagging syntactic arguments of a verb with semantic role labels, a system could be trained to understand fundamental propositional semantics (i.e. who did what to whom, when and how?) using syntactic cues. 

PropBank’s main innovation was in creating a large scale inventory of rolesets (sense disambiguated predicate argument structures) for English verbs, and then having expert human annotators apply them to syntactic parse trees from the Penn TreeBank \citep{taylor2003penn}. The PropBank roleset lexicon consists of verb lemmas organized into frame files. Each frame file contains one or more rolesets representing the different semantic senses associated with the verb, with each roleset providing a predicate label, a written sense definition, and a list of roles corresponding to the semantically-essential participants of the event. PropBank roles are numbered and given short written descriptions rather than more traditional thematic role labels as a way of splitting the difference between semantic and syntactic primacy of the argument. For example, Arg0s correspond to proto-agents \citep{dowty1991thematic}, which also tend to occur as syntactic subjects on verbs, and Arg1s generally correspond to proto-patients, which often occur as syntactic objects. Consider, for example, the 
following roleset for the verb \textit{catch}:

\begin{enumerate}
\setcounter{enumi}{\value{tempcounter}}
\item \label{ex:catch} 
    \textbf{catch-01:} \textit{come into possession of}\\
    \hspace*{5pt} \argline{arg0-}{agent: }{catcher}\\
    \hspace*{5pt} \argline{arg1-}{theme: }{thing caught}\\
    \hspace*{5pt} \argline{arg2-}{source: }{giver}
    \vspace*{0pt}

    `Liu \textbf{caught} the pitch from Cooper'
    \vspace*{3pt}

\setcounter{tempcounter}{\value{enumi}}
\end{enumerate}
    
Annotation consists of two tasks: roleset or sense selection, followed by assigning numbered arguments to denote certain participant roles.  For every instance of a verbal relation in a corpus sentence, annotators would first select a roleset and then tag the nodes in the parse tree governed by the verb with either a) a numbered argument from the roleset, or b) one of a small inventory of general semantic modifier args (ArgMs, e.g., ArgM-LOC (location), ArgM-DIS (discourse markers), ArgM-MNR (manners and instruments)) \citep{bonial2010propbank}. 

One of PropBank's greatest successes was that, across a wide range of corpora and domains, human annotators were able to make these judgments easily and consistently. Inter-annotator agreement (IAA) was consistently high for PropBank -- \citet{bonial2017current} report `exact match' (all constituents and arguments match precisely) IAA for English 
verbal relations at 84.8\%, and `core-arg match' (numbered arguments match and ArgMs match, but 
the specific ArgM, such as Temporal or Locative, need not match) of 88.3\%. 

Propbank 1.0 annotated only verbal relations. During this time, in cases of distinct senses aligning with prepositional verbs (e.g., \textit{``depend on''}) and verb particle Cxns (e.g., \textit{``catch up''}), distinct rolesets were added to the PropBank lexicon to enable accurate annotation of these cases of MWEs. For example: 

\begin{enumerate}
\setcounter{enumi}{\value{tempcounter}}
\item \label{ex:catchB} 
    \textbf{catch\_up-04:} \textit{come even with, to catch up}\\
    \hspace*{5pt} \argline{arg1-}{theme: }{entity in motion}\\
    \hspace*{5pt} \argline{arg2-}{goal: }{goal}
    \vspace*{0pt}

    `The U.S. paper industry needs to \textbf{catch up} with the European industry.'
    \vspace*{3pt}

\setcounter{tempcounter}{\value{enumi}}
\end{enumerate}

Thus, in order to accommodate certain kinds of MWEs, such multi-word relations were stored in the PropBank lexicon.
Notably, this required significant manual effort as the PropBank lexicon of rolesets was driven by annotation.
Thus, in coming upon \textit{``catch up''} for the first time, a PropBank annotator would peruse existing rolesets, and, finding nothing appropriate, would note the instance for a set of PropBank lexicographers 
to create the new roleset, complete with expected argument numbers and a set of examples to guide future annotation. 

Such manual effort becomes untenable with more productive instances of MWEs. The clearest early case of this issue was with light verb constructions (LVCs), mentioned previously. LVCs, when abstracted to the level of, for example \textit{``take''} + deverbal noun, are very productive, but not fully productive. Thus, while many cases of the aforementioned pattern are likely to be instances of LVCs, there are also false positives such as \textit{``\textbf{take a strip} of paper and fold it.''} Furthermore, because the purpose of PropBank is to model the association of semantic roles with particular syntactic patterns, failing to include the MWE roleset precludes assigning the unique roles of the MWE to its unique constituents. As a result, PropBank 1.0 reflected Generative assumptions that there is a single head, specifically a lexical verb, that assigns thematic roles. Thus, the LVC \textit{take a walk}, for example, had previously been annotated with a sense of \textit{``take''} denoting `acquiring something, bringing it with you somewhere.' \textit{``Walk''} was then annotated as the \textsc{Arg1}: \textit{thing taken}. As a result, \textit{``walk''} would be lumped in with concrete objects such as \textit{``book''}, and any semantic roles licensed by the \textit{``walk''} relation would also be mis-tagged as some semantic role of \textit{``take''}. 

To avoid such mistakes and to preclude the necessity of attempting to list productive MWEs in a relatively static lexicon of rolesets, PropBank shifted for the first time to the inclusion of constructional rolesets for LVCs, one type of partially productive MWE. In these constructional rolesets, the fixed, substantive light verb would be tagged with a special `LV' marker, indicating that it provided syntactic scaffolding, but did not contribute a semantic role. Thereafter, the annotators would then leverage the roleset for the event corresponding to the eventive noun, (e.g., \textit{``walk''}), marking that as the relation contributing semantic roles to the surrounding arguments, such as Destinations and Manners. The resulting PropBank annotations of LVCs facilitated modeling the properties of LVCs \citep{bonial2014take}, which in turn provided adequate amounts of high-quality training data to develop the state-of-the-art (at its time) system for detecting LVCs and distinguishing them from identical, non-LVC collocations \citep{chen2015english}. 

The successful modeling of LVCs with a constructional approach was followed by efforts to develop rolesets for other MWEs characterized by fixed and flexible slots. This work coincided with an effort to reorganize the PropBank rolesets according to events with shared semantics and argument structures, as opposed to lexical relations \citep{o2018new}. For example, there was a single roleset reflecting events expressed as the noun or verb \textit{``fear''}, the adjective \textit{``afraid''}, and the LVC \textit{``have fear''}; all instantiations of this event would thus be annotated with the same roleset.  The `unified' PropBank lexicon of rolesets was adopted by both the Abstract Meaning Representation (AMR) \citep{banarescu2013abstract} and the related Uniform Meaning Representation (UMR) projects \citep{van2021designing}. In the AMR project, the notion was to abstract away from syntactic idiosyncracies in order to annotate a more abstract level of meaning. UMR shares this goal, but aims to provide such meaning representations across a wide variety of languages. 

In both of these cases, it is critical to model meaning consistently, despite the array of differences that arise in the Cxns or forms giving rise to that meaning. This included accurately modeling the meaning of MWEs. \citet{bonn2023umr} tackled the challenge of creating rolesets for annotating MWEs across a variety of types and languages. As a result of this work, consider, for example, the following related rolesets for the MWE \textit{``to catch a bug''}, as compared to the lexical relation with a similar meaning, \textit{``contract''}: 

\begin{enumerate}
\setcounter{enumi}{\value{tempcounter}}
\item \label{ex:contract} 
    \textbf{contract-04:} \textit{acquire disease}\\
    \hspace*{5pt} \argline{arg1-}{undergoer: }{sick person}\\
    \hspace*{5pt} \argline{arg2-}{theme: }{disease}\\
    \hspace*{5pt} \argline{arg3-}{source: }{source of disease}
    \vspace*{0pt}

    `Liu \textbf{contracted} giardia from a beaver'
    \vspace*{3pt}

\item \label{ex:catchBug-A}
    \textbf{catch-bug-01:} \textit{become sick}\\
    \hspace*{5pt} \argline{arg1-}{undergoer: }{sick entity}\\
    \hspace*{5pt} \argline{arg2-}{theme: }{virus}\\
    \hspace*{5pt} \argline{arg3-}{source: }{source of contagion}
    \vspace*{0pt}

    `[Liu]\textsubscript{\textsc{arg1}} 
    \textbf{caught}\textsubscript{rel} 
    \textbf{a}
    [stomach]\textsubscript{\textsc{arg2}} 
    \textbf{bug}\textsubscript{rel} 
    [from Cooper]\textsubscript{\textsc{arg3}}'\\
    \hspace*{4pt} \textsc{a} 
    \hspace{27pt} \textsc{b}
    \hspace{26pt} \textsc{c}
    \hspace{2pt} \textsc{d}
    \hspace{50pt} \textsc{e}
    \hspace{2em} \textsc{f}
    
    \textsc{literal:}\\
    \umrA (B / catch-01\\
    \umrB   \textsc{:arg0} (A / \textsc{narg}1-Liu)\\
    \umrB   \textsc{:arg1} (D / \textsc{narg}2-bug\\
    \umrC       :mod (C / stomach))\\
    \umrB   \textsc{:arg2} (E / \textsc{narg}3-Cooper))
    \vspace*{0pt}
    
    \textsc{idiomatic:}\\
    \umrA (B / contract-04\\
    \umrB   \textsc{:arg1} (A / \textsc{Narg}1-Liu)\\
    \umrB   \textsc{:arg2} (DE / \textsc{Narg}2-norovirus)\\
    \umrB   \textsc{:arg3} (F / \textsc{Narg}3-Cooper))
    \vspace*{3pt}

\item \label{ex:interest}
    \textbf{interest-01:} \textit{provoke/exhibit curiosity}\\
    \hspace*{5pt} \argline{arg0-}{stimulus: }{subject of interest}\\
    \hspace*{5pt} \argline{arg1-}{experiencer: }{interested entity}\\
    \hspace*{5pt} \argline{arg2-}{instrument: }{further descr. of ARG0}
    \vspace*{3pt}
    
\item \label{ex:catchBug-B}
    \textbf{catch-bug-02:} \textit{become fervently interested in}\\
    \hspace*{5pt} \argline{arg1-}{experiencer: }{interested entity}\\
    \hspace*{5pt} \argline{arg2-}{stimulus: }{subject of interest}
    \vspace*{0pt}
    
    `[Liu]\textsubscript{\textsc{arg1}} 
    \textbf{caught}\textsubscript{rel} 
    \textbf{the} 
    [MWE]\textsubscript{\textsc{arg2}} 
    \textbf{bug}\textsubscript{rel}'\\
    \hspace*{4pt} \textsc{a}
    \hspace{27pt} \textsc{b}
    \hspace{30pt} \textsc{c}
    \hspace{7pt} \textsc{d}
    \hspace{35pt} \textsc{e}

    \textsc{literal:}\\
    \umrA (B / catch-01\\
    \umrB   \textsc{:arg0} (A / \textsc{narg}1-Liu)\\
    \umrB   \textsc{:arg1} (E / \textsc{narg}2-bug\\
    \umrC       :mod (D / MWE)))
    \vspace*{0pt}
    
    \textsc{idiomatic:}\\
    \umrA (B / interest-01\\
    \umrB   \textsc{:arg0} (D / \textsc{narg}2-MWEs\\
    \umrB   \textsc{:arg2} (A / \textsc{narg}1-Liu)\\
    \umrB   :degree intensifier)
\setcounter{tempcounter}{\value{enumi}}
\end{enumerate}

\noindent Rolesets for idiomatic MWEs in UMR are quite descriptive about the relationships between the MWE’s tokens and the elements in the literal and figurative frames, as illustrated in Examples~\ref{ex:catchBug-A} and \ref{ex:catchBug-B}.
First, numbered roles are provided for participants in the idiomatic frame as well as any modifiers the expression can take. In Example~\ref{ex:catchBug-B}, the Arg1 (experiencer) of the MWE roleset catch-bug-02 corresponds to the Arg0 of the literal roleset catch-01 and Arg2 of the idiomatic roleset interest-01 (these mappings are denoted in each argument by the nested ``\textsc{narg1}-Liu''). Additionally, the expression \textit{``catch a bug''} is modifiable, i.e. there is a schematic slot that can be optionally filled that modifies \textit{``bug''} (e.g., \textit{stomach} bug). To account for this, both idiomatic rolesets provide slots with additional semantic information about the character of the bug, such as the type of virus or the subject of interest. 

By modeling MWEs using such rolesets, several types of information are gained. First, the meaning pole: the overall meaning of the MWE, as well as its similarity to the meaning of related expressions (e.g., \textit{``contract''}) is spelled out. Second, the form pole: the roleset spells out any fixed words in the expression (e.g., \textit{``catch bug''}), while also noting the flexible slots in the expression and supplying a semantic role designation for annotating the flexible slot. As a result, the finished annotations are able to model the syntactic properties of the flexible slots as well as lexical-semantic information about what kinds of words are ``attracted'' to the slot. Thus, MWEs annotated and modeled according to this constructional paradigm enable the detection of such MWEs and distinction of these cases from non-idiomatic, identical collocates. What's more, such modeling facilitates a deeper semantic understanding of the structures (in particular, the semantic roles of flexible slots) than approaches that model MWEs based on collocational strength alone. In the next section, we extend this approach to modeling morphologically complex Cxns, which we argue to be a similar challenge to modeling MWEs. 

\section{Accounting for MWEs in Morphologically Complex Languages}
\label{sec:MWEs-MCLs}

\noindent Croft says: ``Morphosyntax is the analysis of the internal structure of utterances, both above the word level and below it. Why combine morphology and syntax? Because grammatical constructions involve both'' \parencite*[3]{croft2022morphosyntax}. In this section, we look at how a CxG approach that redefines the `word’ in `multiword expression' as a \textit{meaningful morphosyntactic unit (MMU)} (rather than purely syntactic unit) can help make sense of MWE phenomena, including idioms, in polysynthetic languages. We use Arapaho, a highly polysynthetic and agglutinating Algonquian language of the North American plains, as an illustration.

\subsection{MWEs and the Notion of `Wordhood'}
\label{subsec:wordhood}

\noindent Section \ref{subsec:peripheral} noted that “the machinery needed for describing the so-called minor or peripheral constructions will have to be powerful enough to be generalized to more familiar structures.” Approaching MWEs in morphologically complex languages as \textit{multi-MMU} Cxns allows us to do two important things: 1) to include polysynthetic languages in discussions of MWEs, and 2) to place MWE phenomena in the context of the more general diachronic phenomena of word-building, lexicalization, and development of the mental constructicon. Far from being ‘peripheral’, MWEs are in fact illustrative of the very heart of language development. 

The process of MWE formation is very much like the process of lexicalization in general. Lexicalization proceeds as follows: through usage over time, conceptually independent meaning units (e.g., words or other MMUs) that were previously instantiated in a variety of higher-order Cxns (e.g., morphemes being combined into words in the case of morphological Cxns, or words being combined into phrases in the case of lexical Cxns) begin to be used at a high frequency in one particular instantiation, which entrenches the forms within that higher-order Cxn. As a result, what was previously readily analyzed as consisting of several sub-Cxns begins to merge into a single coherent semantic concept that is somehow more than the sum of the parts. As this occurs, that new concept takes its place as a new entry in the mental constructicon. This occurs not only at the level of an individual's mental constructicon, but also in the common constructicon shared by the community of speakers. While word-building through inflectional and derivational processes is not generally thought of as lexicalization, the two are closely related. \citet{borin2021multiword} says that derived forms frequently end up becoming associated with a narrower semantic interpretation than their source forms. In his example from Finnish, the denominal/deadjectival nominalizing suffix \textit{``-sto''/``-stö''} functions productively and compositionally to mean `a collection of X'. However, it is not uncommon for the derived forms to take on much more specific meanings, such as \textit{``kirjasto''} (`a collection of books') coming to refer to a `library' \parencite*[225]{borin2021multiword}. The compositional usage is derivation, and the semantically shifted usage is lexicalization. Derivation does not necessarily lead to lexicalization, but it does feed it. 

MWEs can be considered a special form of lexicalization insofar as they represent the consolidation of formerly conceptually-independent meaning units into a single lexical sense. Some MWEs may eventually turn into a single orthographic or phonological word, as has occurred with \textit{``nevertheless''}, and \textit{``insofar''}--although clearly this is more likely to occur with fully substantive and contiguous MWEs, and not at all likely with schematic MWEs. Still, the term `multiword expression' itself betrays the position held in much of the literature that MWEs are a phenomenon that occurs above the word level, not below. 

But are MWE phenomena relevant to languages in which the definition of `word' is much less likely to refer to something that aligns with a single semantic concept? The absence of inclusion of polysynthetic languages (and other examples of morphological complexity such as noun incorporation) in so much of the MWE literature would seem to suggest that the most important criterion for MWE-hood is the involvement of multiple `words' in the orthographic (white space-delineated) or phonological sense. We argue that white space delineation is not a particularly relevant criterion for MWE-hood in polysynthetic languages \parencite{dixon2002word}, and that idiomaticity is a far better one. Cxns that bear the hallmarks of lexical idiosyncrasy and (morpho)syntactic creativity are found in polysynthetic languages just as they are found in isolating and fusional languages, and the forces that drive these phenomena appear to be like in kind. 

Although they only circle the issue, \citeauthor{baldwin2010multiword}'s definition of MWE \textit{does} make room for inclusion of `lexical items' that comprise a single orthographic or phonological word, as long as they ``can be decomposed into multiple lexemes'' and ``display lexical, syntactic, semantic, pragmatic, and/or statistical idiomaticity'' \parencite[269]{baldwin2010multiword}. We follow \citeauthor{borin2021multiword} \parencite*[228]{borin2021multiword} in interpreting `lexeme' and `lexical item' here to be synonymous, and we also propose that these terms are equivalent to our MMUs.

Viewing MWEs through a CxG lens makes it possible to include idioms in morphologically complex languages in the conversation. Furthermore, constructional templates used for modeling MWEs turn out to be equally useful for symbolic computational modeling of morphologically complex predicates, whether idiomatic or not. This serves an important need, as there is an increasing desire for symbolic representations of languages with fewer computational resources, and polysynthetic languages overwhelmingly belong to this category.

\subsection{Words and Other Constructions in Arapaho}
\label{subsec:Arapaho}

\noindent Arapaho is a highly polysynthetic and agglutinating Algonquian language of the North American Great Plains. Although it was once spoken widely in Colorado, Wyoming, Texas, Oklahoma, New Mexico, Montana, and up into Alberta and Saskatchewan, it is now critically endangered with fewer than 150 speakers remaining on the Wind River Reservation in Wyoming. Although endangered, the Arapaho Language Project \footnote{https://verbs.colorado.edu/ArapahoLanguageProject/} has compiled significant resources over the past 20 years including a reference grammar \parencite{cowell2011arapaho}, an online Arapaho Lexical Database with around 29,000 active entries \parencite{cowell2021textDB}, and a conversational database of spoken Arapaho in text and video format \parencite{cowell2010conversational, cowell2024conversational}. 

\begin{figure}
\includegraphics[width=1.0\textwidth]{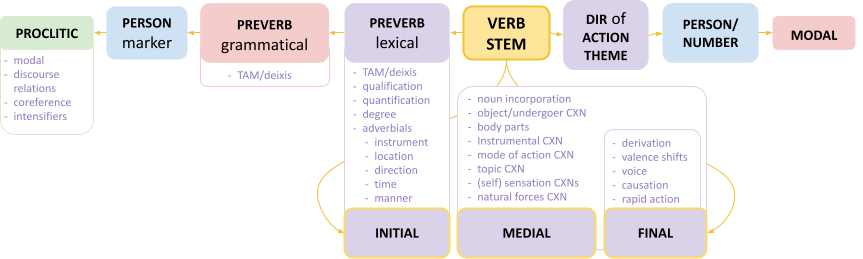} 
\captionof{figure}{Constructional template for Arapaho verbs, listing part of speech and semantic content types for each slot in the structure.}
\label{fig:verbSlots}
\end{figure}

Structurally, the language places enormous importance on the verb,\footnote{We define verb in a language-internal fashion \parencite{croft2023word}; thus, `verb' is shorthand for the Arapaho \textsc{verb} Cxn.} using extensive head-marking morphology on verb stems and encoding as much information as possible through `verb-words’ (words rooted in verb stems, which may function syntactically as other parts of speech such as deverbal nouns). A single verb-word may express an entire complex proposition, with all participants, adverbials, and lexical (i.e., semantically contentful) modifiers expressed via affixes or incorporated nouns \parencite{cowell2011arapaho}. Verb-words adhere to a strict internal template structure that organizes their morphosyntactic components, including their argument structures. Outside of the verb-word, Arapaho’s word order is almost entirely free. Figure \ref{fig:verbSlots} presents the constructional template for the \textsc{verb-word} Cxn in Arapaho \parencite{bonn2025problem}.

In stark contrast to the high frequency of verbal incorporation, Arapaho exhibits a countering force of \textit{deincorporation} in which pragmatically salient information is emphasized by being extracted from the verb and placed independently elsewhere in the sentence \parencite[9]{cowell2011arapaho}. Deincorporation is truly a pragmatic endeavor; often, the semantic arguments that are realized as independent noun phrases are either a) already lexically entailed by the verb stem (meaning they are, semantically-speaking, redundant), or b) not arguments syntactically licensed by the Arapaho \textsc{Transitive} Cxn (meaning they are not the grammatical subject or object). Consider the Arapaho sentence in \ref{horseNotBuff}.
\footnote{Orthographic conventions in Arapaho: the number `3` for a voiceless dental fricative /$\theta$/; an apostrophe for a glottal stop /?/} 
\footnote{Glossing conventions in Arapaho: \textsc{na} = animate noun; \textsc{obl.poss} = obligatorily-possessed noun; \textsc{ic} = initial change; \textsc{detach} = detachment Cxn; 12 = pronominal index, 1st and 2nd person together; 3 = pronominal index, 3rd person; 4/3 = pronominal index, obviative 3rd person (4) as subject with a proximal 3rd person (3) as direct object}
\begin{exe}
\ex\label{horseNotBuff}
  \gll Noh \textbf{woxhooxeb-ii} ne'-nih-ii'-iiniikohei-no', \\
       and \textbf{horse-\textsc{pl}} that-\textsc{past}-when-ride.a.horse.around-12 \\
  \gll hoow-uuhu' heneecee-no'.\\
       3.\textsc{neg}-\textsc{adv} buffalo.bull-\textsc{na.pl}\\
  \glt `And that is why we rode horses around instead of buffalos.' \parencite{cowell2021textDB}
\end{exe}

\noindent This sentence shows a morphologically complex verb-word with an independent noun phrase and an independent adverbial phrase modifying it externally. It is important to know that including \textit{``woxhooxebii''} `horses' explicitly as an independent noun phrase is not required, as `horses' are already encoded (in this case lexically entailed) in the verb-word. From a Generative, lexicalist perspective, since the verb stem is grammatically intransitive, this argument is not grammatically licensed by it. Instead, it is licensed by a specialized Arapaho Cxn called the \textsc{detachment} Cxn \parencite[231]{cowell2011arapaho}: the presence of \textit{``woxhooxebii''} as an independent noun is an example of deicorporation of a pragmatically salient participant in order to emphasize it (`horses, not buffalos'). 

\subsubsection{Nouns and Noun Incorporation in Arapaho}
\label{subsubsec:nouns}

\noindent \textit{``Wóxhoox(ew)''} (\textsc{na}) in \ref{horseNotBuff} is an independent noun form in Arapaho, meaning it does not need to be marked with any additional morphology to exist as its own orthographic word. Many nouns in Arapaho also have an alternate obligatorily possessed form, which can also be used as orthographic words, but must be prefixed by a pronominal index for the possessor. Interestingly, the obligatorily possessed form for `horse’, \textit{``-toníhi’''}  (\textsc{na.obl.poss}), literally means `X’s \textit{pet}’ (\textit{``nó-toníhi’''} `my horse’ and  \textit{``hó-toníhi’''} `your horse’). The `pet’ sense is still productive and can be used with a more specific animal noun, as in  \textit{``nótoníhi’ beníixóxko’ó’''} `my pet goat’ \parencite*[67]{cowell2011arapaho}. In contrast to \citeauthor{borin2021multiword}’s Finnish example of `a collection of books’ lexicalizing to ‘library’, \textit{``-toníhi’''} `horse’ illustrates semantic shift within a single morpheme, but not lexicalization, since the semantic shift did not result in a loss of compositionality of the component forms.

Independent possessed nouns can be derived into adverbial prefixes that can be added to verb-words outside of the stem (e.g., \textit{``wóxhooxu-''}). Their etymological roots often appear as incorporated forms inside of verb stems (e.g., \textit``{-ôoxew-''} →  \textit{``bénoh<óoxeb>éí-''} (\textsc{vai.incorp}), `to water horses or other domestic animals’). Obligatorily possessed forms, being relational, participate in the denominalizing \textsc{predicative possession} Cxn, which productively turns them into verb stems, as in \textit{``hitonihi'-''} (\textsc{vai.incorp}) `own a horse’, ‘have something as a pet’.

Generally speaking, incorporated nouns inside of verb stems can be considered to have lexicalized into the stem, meaning that speakers already consider them to be a single conceptual unit. For our purposes of identifying MWEs within \textsc{verb words}, we consider these stems to already be single MMUs and thus not examples of MWE-hood. The exceptions to this are schematic verb-stem forming Cxns like \textsc{predicative possession} and others found in Table \ref{tab:NITypes} below that productively take a noun in one of their slots, and whose morphosyntactic behavior is projected by the schematic Cxn rather than by the combination of the parts.

\subsubsection{Verb Stem Constructional Templates}
\label{subsubsec:stemCxns}

\noindent Verb stems in Arapaho are internally complex, consisting of multiple morphologically analyzable (but not necessarily morphologically decomposable) initials, medials and finals. In \ref{horseNotBuff}, the stem \textit{iiniikohei-} `ride a horse around' is an instantiation of the \textsc{mode of action} Cxn. This is a fully schematic Cxn which pairs an action concrete-final (\textit{-iikohei} `riding') with a manner-adverbial initial (\textit{iin-} `around'). Since this and the other noun incorporation Cxns listed in Table \ref{tab:NITypes} are extremely productive and compositional, the verb stems they produce are not considered lexicalized unless further semantic shift occurs, and the elements that fill the slots of these Cxn templates may be considered MMUs. 

In \textit{``iiniikohei-''}, the concrete action final \textit{``iikohei''} itself can be analyzed further. It includes a derivational final marking it as an animate intransitive verb (AI), \textit{``\mbox{-yei}''}, which has been applied to the lexically substantive action final \textit{``iiko''} `go by horse’. Even though \textit{``iiko''} is still semantically complex -- it encodes both an event concept and a participant concept -- the form is fusional and does not appear to be able to be broken down further into distinct morphemes.

\begin{singlespace}
\begin{longtable}{p{0.26\textwidth} p{0.74\textwidth}}
\caption{Productive noun Incorporation constructional templates in Arapaho \parencite{cowell2011arapaho}}. \\
\label{tab:NITypes} \\
\toprule
    \multicolumn{2}{l}{\tabSC{Noun Incorporation Cxns for Verb Stem Formation}} \\
\midrule
\endfirsthead
\toprule
    \multicolumn{2}{r}{\tabSC{Noun Incorporation Cxns for Verb Stem Formation}} \\
\midrule
\endhead
\midrule
    \multicolumn{2}{r}{\textit{Continued on next page}} \\
\endfoot
\bottomrule
\endlastfoot

\textbf{\textsc{mode of action}} & F<A\textsubscript{transitive-action-root-\textsc{initial}}--B\textsubscript{manner-\textsc{medial/concrete-final}}> \\
 & \textit{iin-iikoh<ei>-} `ride a horse around' \\
 & around\textsubscript{\textsc{a}}--ride.horse<AI-final>\textsubscript{\textsc{b}} \\
\whitecmidrule

\textbf{\textsc{instrumental}} & F<A\textsubscript{transitive-action-root-\textsc{initial}}--B\textsubscript{implement/method-\textsc{concrete-final}}> \\
 & \textit{to’-oxon-} `kick’ \\
 & hit\textsubscript{\textsc{a}}--foot\textsubscript{\textsc{b}} \\
\whitecmidrule

\textbf{\textsc{patient/undergoer}} & F<A\textsubscript{action-root-\textsc{initial}}--B\textsubscript{patient-\textsc{medial/concrete-final}}> \\
 & \textit{bii--bin<ee>-} `eat berries’ \\
 & eat\textsubscript{\textsc{a}}--berries<AI-final>\textsubscript{\textsc{b}} \\
\whitecmidrule

\textbf{\textsc{body part}} & F<A\textsubscript{condition/characteristic-root-\textsc{initial}}--B\textsubscript{body-part-\textsc{medial}}> \\
 & \textit{nooku--3e’en<ee>-} `have a white wing’ \\
 & white\textsubscript{\textsc{a}}--wing<AI-final>\textsubscript{\textsc{b}} \\
\whitecmidrule

\textbf{\textsc{topic}} & F<A\textsubscript{condition/characteristic-root-\textsc{initial}}--B\textsubscript{natural-object/phenomenon-\textsc{concrete-final}}> \\
 & \textit{tooyo3-eese-} `the wind is cold’ \\
 & cold\textsubscript{\textsc{a}}--wind\textsubscript{\textsc{b}} \\
\whitecmidrule

\textbf{\textsc{sensation}} & F<A\textsubscript{experiential-quality-root-\textsc{initial}}--B\textsubscript{sensory-modality-\textsc{concrete-final}}> \\
 & \textit{nii’-noeyoti-} `it looks good’ \\
 & good\textsubscript{\textsc{a}}--visual-appearance\textsubscript{\textsc{b}} \\
\whitecmidrule

\whitecmidrule
\end{longtable}
\end{singlespace}

\subsubsection{The Flexibility of Arapaho Words}
\label{}

\noindent `Words’ in Arapaho may be defined in multiple ways \parencite{dixon2002word}. \textit{Orthographic words}, \textit{syntactic words}, and \textit{phonological words} pick out the same set. These are the chunks that can move around freely in a sentence’s syntactic structure. When Arapaho is said to have a free word-order, these are the `words’ that are meant. These are generally identifiable because of phonological processes that apply only at word boundaries (white space, orthographically speaking), such as the application of /h-/ to any vowel-initial candidates. 

However, examples of deincorporation show that there is more to the story, as there are certain morphophonological processes that apply to verb-words only once, even if the verb-word has been split into multiple phonological/syntactic words using the \textsc{detachment} Cxn.  

\begin{exe}
\ex\label{scout}
  \gll wootii yoo3i-ini iiyoo3i-ini notikoni-3i' \\
       like \textsc{ic}.hidden-\textsc{detach} clean-\textsc{detach} scout-\textsc{3pl} \\
  \glt `They would scout while staying hidden, in the proper, clean way' \parencite[232]{cowell2011arapaho}
\end{exe}

\noindent In \ref{scout}, \textit{``yoo3i''} and \textit{``iiyoo3i''} are both prefixes that have been detached. While they are labeled as particles in the glossing, Cowell and Moss Sr. indicate that they fall into a grey area. A morphophonological process called `initial change' is supposed to apply to any verb that is not otherwise marked for tense or aspect, infixing an /-en-/ or /-on-/ in the first element of the verb (when the first vowel is a long vowel, and depending on vowel harmony). Although \textit{``yonoo3ííni''} and \textit{``hííyoo3iini''} are independent words orthographically, syntactically, and phonologically (note that \textit{``hííyoo3iini''} has received a word-initial /h-/), they must still be morphologically dependent on the verb because the initial change has stayed attached to the deincorporated \textit{``y\textbf{on}oo3ííni''} rather than being moved to the chunk now labeled as the verb. So the designation of detached prefixes as particles comes from the perspective that they are \textit{syntactic words}, although morphosemantically and constructionally speaking, they are still part of the verb-word. This demonstrates that the \textsc{detachment} Cxn produces noncontiguous verbal MWEs in the canonical orthographic sense (much as verb-particle Cxns are considered MWEs) although these are not MWEs that we expect to become idiomatic in Arapaho.

The fact that Arapaho allows meaningful elements to be deincorporated from the verb at all demonstrates that speakers view them as semantically independent. These elements are \textit{specifically} elements that fill slots in the \textsc{detachment} Cxn template and participate in its propositional argument structure -- and this is the definition of `meaningful morphosyntactic unit’, which we propose as the appropriate unit to use when discussing MWEs in Arapaho and other polysynthetic languages.

At this point it is useful to distinguish several concepts. \textbf{Word-building} is the process of constructing the orthographic/phonological words of Arapaho. The first step of this is filling the slots of word-form Cxns, such as the \textsc{verb word} Cxn or the \textsc{obligatory possessed noun} Cxn, although the \textsc{detachment} Cxn (another word-building Cxn) should also be applied as part of the process to ensure that phonological rules can be applied correctly. Note that word-building and proposition-building are closely related in Arapaho and other polysynthetic languages, and this is why identifying MWEs has been so difficult. Proposition-building \textit{also} begins with filling in the slots of the \textsc{verb-word} Cxn in Arapaho. 

\textbf{Lexicalization} is not necessarily tied to word-building in Arapaho. For example, the output of proposition-building processes should not be considered lexicalized as long as compositionality has been retained. If compositionality is lost and meaning becomes more than the sum of the parts, lexicalization has occurred. 

\textbf{MWE-formation} may occur if a single word-form constructional template results in multiple orthographic forms (like with the output of the \textbf{verb word} Cxn + \textbf{detachment} Cxn), or if lexicalization occurs across two or more MMUs. This includes cases in which the MMUs lie within the same word-form constructional template as well as cases where the two MMUs are separate orthographic/phonological words. 

It should be noted that in cases where an MWE develops within elements of a verb-word, it is not necessarily the case that the entire orthographic word becomes the MWE. For example, if a discourse relation from the first (proclitic) slot in the template lexicalizes with a tense or aspect marker from the third (grammatical preverb), those two elements form an MWE without involving any of the rest of the propositional elements. A caveat here is that the two MMUs should still be identifiable to speakers as having individualized interpretations as well as a combined interpretation.

Overall, Arapaho provides evidence that the (orthographic)
‘word’ is not the most salient unit of the language. This aligns with a CxG perspective, which posits that Cxns are the fundamental unit of a language,
and that the constructicon is made up of a taxonomically related set of Cxns that are characterized by differing levels of hierarchical abstraction. As a result, Cxns are often characterized as existing at the morphological level, the word level and the phrase level, depending upon the form-meaning pairing under analysis. However, it should be noted that this characterization may draw upon an unwarranted assumption -- that there is a clear distinction between these levels in the language. Arapaho provides counter-evidence to this problematic assumption. Just as \citet{croft2022morphosyntax} posits that parts of speech can only be defined according to the Cxns of a language, we posit that the formal units of a language may not readily align with notions of morpheme and lexeme, and instead should also be defined according to the Cxns of the language, with special attention paid to where we see formally identical Cxns instantiated both independently and within higher-order Cxns.

\subsection{Rolesets for Arapaho MWEs}
\label{subsec:ArapahoRolesets}

\noindent In this section, we shift again to a real-world use case that benefits from leveraging a constructional modeling of MWEs, but this time in Arapaho. Here we focus on the UMR annotation of Arapaho. UMR inherits the PropBank/AMR shared style of representing the lexicon in the form of rolesets, which then guide the assignment of semantic roles, designated as `Arg' numbers (see Section~\ref{sec:EnglishPropBank} for details in English). Like AMR, UMR represents sentential meaning as a directed, acyclic graph, but UMR includes flexibility as to what is included in this graph. This flexibility facilitates a language-dependent, data-driven representation of meaning in a particular language.  Without such flexibility (as is the case with AMR), the meaning representation tends to reflect the idiosyncrasies of the one or two languages that it  was originally developed to handle (English and Chinese, in the case of AMR) \citep{banarescu2013abstract}.

When an MWE develops in a verb-word that scopes over the predicate, a new predicate is added to the mental constructicon with a unique argument structure. These are the cases where idiomatic rolesets may be created. Example \ref{ghost-arrow} presents an idiomatic MWE, synonymous with `catch a bug' from Example~\ref{ex:catchBug-A}. 

\begin{enumerate}
\setcounter{enumi}{\value{tempcounter}}
    \item \label{ghost-arrow}
    \gll nih-3iikon-ceb-eit \\
         \textsc{past}-ghost-shoot.with.arrow-4/3 \\
    \glt \textit{lit.:} `a ghost shot him with an arrow.'
    \glt `He caught a disease.'\\
    \hfill\citep{bonn-etal-2023-umr}
\setcounter{tempcounter}{\value{enumi}}
\end{enumerate}

\noindent In the morphological breakdown, ``\textit{3iikon-}'' is a nominal preverb that refers to the ghost, and ``\textit{ceb-}'' is a verb stem that lexically entails shooting with an arrow. The ‘ghost’ indexes to the subject argument, marked as 4th person with the pronominal suffix ``\textit{-eit}''. Being 4th person (the code used for obviative pronominals in Arapaho transcriptions) makes this entity indefinite, which means their identity may not be known, or they are at least not the primary active entity in discourse. In effect, the focus is actually on the direct object of the verb, the 3rd person, \textit{getting} sick, rather than the act of communicating the disease or the identity of the contagious party.  

Work is currently underway to create rolesets for a selection of the Arapaho lexicon, including this one \citep{bonn2025problem}. Like MWE rolesets for English, this roleset includes a metaphorical mapping between literal and figurative meanings, as in \ref{ex:3iikonceb}. While the designation of MMUs is helpful in identifying MWEs in polysynthetic languages, rolesets provide deeper semantic representations that include smaller morphological units (extracted from verb stems) and even implicit but semantically essential arguments, such as the `arrow’, which is lexically entailed by the verb. 

\begin{enumerate}
\setcounter{enumi}{\value{tempcounter}}
    \item \label{ex:ceb}
    \textbf{ceb-01:} \textit{shoot}\\
    \hspace*{5pt} \argline{arg0-}{agent: }{shooter}\\
    \hspace*{5pt} \argline{arg1-}{patient: }{victim}\\
    \hspace*{5pt} \argline{arg2-}{instrument: }{arrow (semi-fixed)}
\setcounter{tempcounter}{\value{enumi}}
\end{enumerate}

\begin{enumerate}
\setcounter{enumi}{\value{tempcounter}}
    \item \label{ex:3iikonceb}
    \textbf{3iikonceb-01:} \textit{(cause to) become sick}\\
    \hspace*{5pt} \argline{arg0-}{causer: }{contagious entity}\\
    \hspace*{5pt} \argline{arg1-}{experiencer: }{sick entity}\\
    \hspace*{5pt} \argline{arg2-}{stimulus: }{illness}
    \vspace*{0pt}

    `[nih]\textsubscript{\textsc{infl}}-
    3iikon\textsubscript{\textsc{arg0}}-
    ceb\textsubscript{\textsc{rel}}
    -[eit]\textsubscript{\textsc{arg0/arg1}}\\
    \hspace*{4pt} \textsc{a} 
    \hspace{27pt} \textsc{b}
    \hspace{27pt} \textsc{c}
    \hspace{27pt} \textsc{d e}
    
    \textsc{literal:}\\
    \umrA (C / ceb-01\\
    \umrB   \textsc{:arg0} (BD / \textsc{narg}0-3iikon\\
    \umrC       :refer-definiteness obviative)\\
    \umrB   \textsc{:arg1} (E / \textsc{narg}1-3S)\\
    \umrB   \textsc{:arg2} (C / \textsc{narg}2-<arrow>))
    \vspace*{0pt}
    
    \textsc{idiomatic:}\\
    \umrA (C / catch-bug-01\\
    \umrB   \textsc{:arg3} (BD / \textsc{Narg}0-ghost)\\
    \umrB   \textsc{:arg1} (E / \textsc{Narg}2-illness)\\
    \umrB   \textsc{:arg2} (F / \textsc{Narg}3-3S))
\setcounter{tempcounter}{\value{enumi}}
\end{enumerate}

\noindent The roleset for the idiomatic expression contains three arguments for the contagious party, the infected party, and the illness; the roleset for the literal verb `ceb' contains three arguments for the ghost, the victim, and an arrow. The constructional template shows that \textit{3iikon} `ghost’ and \textit{ceb} ‘shoot with arrow’ are the MMUs that have become fixed, while other slots in the \textsc{verb word} constructional template remain schematic. Although it is only an implicit argument morphosyntactically speaking, the arrow is a critical piece of the metaphorical mapping between literal and figurative interpretations of this expression, since it maps to the disease.

\subsubsection{Extending MWE rolesets to Morphologically Complex Predicates}
\label{subsubsec:MCP}

\noindent The development of MWE rolesets has been critical to the development of rolesets for polysynthetic languages. The core innovation of MWE rolesets was the inclusion of the token-slot mappings, and these have been used to model token-slot mappings in general cases of morphological complexity even when idiomatic meaning is not present-- which as we have seen is the case with the majority of Arapaho verb-words. These rolesets make it possible to provide deeper semantic entailments that define semantically essential morphology below the word or verb stem level, as in \ref{ex:iiniikohei}:

\begin{enumerate}
\setcounter{enumi}{\value{tempcounter}}
\item \label{ex:iiniikohei}
    \textbf{iiniikohei-01:} \textit{ride a horse around}\\
    \hspace*{5pt} \argline{arg0-}{agent: }{rider}\\
    \hspace*{5pt} \argline{arg1-}{theme: }{horse}\\
    \hspace*{5pt} \argline{arg2-}{path: }{place ridden}
    \vspace*{0pt}

    [`iin]\textsubscript{\textsc{arg2}-around}-
    iikohei\textsubscript{\textsc{rel}-go.by.horse}
    -[\textsc{infl}]\textsubscript{\textsc{arg0}}\\
    \hspace*{4pt} \textsc{a} 
    \hspace{27pt} \textsc{b}
    \hspace{27pt} \textsc{c}
    
    \textsc{semantic representation:}\\
    \umrA (AB / iiniikohei-01\\
    \umrB   \textsc{:arg0} (C / \textsc{narg}0-[rider])\\
    \umrB   \textsc{:arg1} (B / \textsc{narg}1-<horse>)\\
    \umrB   \textsc{:arg2} (A / \textsc{narg}2-[place ridden]))
    \vspace*{0pt}
\end{enumerate}

\noindent This representation allows semantically essential elements included as part of the verb stem or lexically entailed (yet implicit) to be mapped to the numbered semantic arguments of the roleset as well as to structures within the UMR-format semantic representation.

Since polysynthetic languages tend to also be low-resource languages, speaker communities who wish to participate in development of NLP resources for these languages cannot expect to rely on state-of-the-art AI strategies like LLMs -- there is simply nowhere near enough training material to even begin to create such a thing. The Arapaho Conversational Database contains 29,000 sentences, as opposed to the billions upon billions needed to feed an LLM. 

There is considerable worry that NLP resources for low-resource languages that rely on brute force methods will end up creating faulty representations of the language, causing irreparable harm to documentation of the language and especially to the language communities themselves. Once the bad-NLP horse has left the barn, so to speak, it will be virtually impossible to put it back in. Symbolic meaning representations like PropBank and UMR are well established as highly successful with smaller training sets, and continue to be increasingly relevant to smaller language communities that wish to participate in NLP research.


\section{Critical Aspects of MWE Research in the Era of LLMs}
\label{sec:LLMs}
In the previous sections, we have provided a description of how we can build symbolic resources, grounded in CxG theory, supporting automatic recognition and interpretation of MWEs. However, in English and other resource-rich languages, we recognize that subsymbolic methodologies for a variety of NLP tasks requiring MWE analysis are dominant. Thus, in this section, we turn to how LLMs handle MWEs.  

LLMs, primarily trained on next-token prediction, have demonstrated an impressive range of capabilities. To dismiss them merely as statistical next-word prediction tools would be an underestimation of their potential, as demonstrated by numerous recent studies. However, assuming these models are all-powerful and capable of complex `reasoning' similarly misrepresents their limitations~\citep{lu-etal-2024-emergent}. The truth likely lies somewhere in between. In this section, we provide a focused, brief overview of what LLMs can reliably achieve, what they are less consistent at accomplishing, and the implications of these boundaries for ongoing research into MWEs. 
Based on this discussion, we design targeted experiments to expose specific strengths ad weaknesses in LLMs, highlighting the critical areas in MWE and CxG research that remain essential -- perhaps even more so -- in light of the rise of LLMs.

In an attempt to first differentiate the aspects of language that LLMs can handle well from those they handle less well we begin by providing some clear definitions. Specifically, we adopt and refine the definitions of \emph{formal linguistic competence} and \emph{functional linguistic competence} from ~\citet{MAHOWALD2024517}, who define formal linguistic competence as the abilities that typically encompass those required to produce and comprehend language, while functional abilities pertain to using language to achieve specific goals. However, for our purposes, this distinction needs to be further refined to effectively pinpoint the challenges LLMs face with regard to MWEs, and language `understanding' in general.

\subsection{Formal and Functional Linguistic Competence}
We define formal linguistic competence as the set of capabilities required to comply with the structural rules and patterns that govern a language. We explicitly exclude the notion of `understanding,' present in the original definition, and our focus is purely on the phonological, morphological, structural and lexical semantic aspects of language, which typically are in line  with components of the traditional NLP pipeline. Specifically, these include:
\begin{itemize}
    \item Phonological and morphological knowledge: Permissible sound combinations and productive morpheme usage.
    \item Syntactic knowledge: Sentence structure and agreement rules.
    \item Lexical knowledge: Ability to associate words with their correct grammatical and contextual roles.
    \item Constructional sensitivity: Ability to recognize linguistic regularities, exceptions, and idiomatic expressions, including MWEs.
\end{itemize}

A central feature of formal linguistic abilities is their \emph{variability across languages and the need for expert training to explicitly label or annotate them.} Native speakers, while possibly not being able to annotate or explicitly identify these elements, can typically use them fluently and without formal training, relying on implicit knowledge, which we will call tacit linguistic knowledge (see Section \ref{sec:meta-tacit} below for a contrast to meta-linguistic information).

For example, consider the sentence: \textit{``The house key is on the cabinet.''} Here, the subject (\textit{``key''}) and the verb (\textit{``is''}) agree in number. This agreement must remain consistent even when the subject and verb are separated by a lengthy relative clause: \textit{``The house key that is brown and was lost last winter only to be found again is on the cabinet.''} Despite this additional complexity, speakers instinctively ensure subject-verb agreement based on the noun (\textit{``key''}). This ability of native speakers stems from implicit familiarity with the relevant Cxns, including the \textsc{intransitive} as well as the \textsc{noun-phrase} and \textsc{relative-clause} Cxns.\footnote{Subject-verb agreement, or concord, is sometimes modeled as a constraint in CxG, as opposed to a Cxn, because there is no clear meaning pole, e.g. \citet{hoffmann_2022}; however, \citet{croft2001radical} argues, and we tend to agree, that \textit{-s} does have a semantic function in allowing for the identification of roles for speakers even when encountering unfamiliar words.}
Therefore, while native speakers can track and use these structures effortlessly, they are not likely to be able to complete tasks such as producing parse trees, which require specialized training. 

There is another important feature of formal linguistic structures: they are theoretical constructs developed by linguists -- they do not exist as explicit and tangible entities within a speaker’s mental representation of language. Therefore, they serve as tools to analyze the implicit structural knowledge that underpins fluent language use. These tools are \emph{post hoc} analytical frameworks, and capture aspects of language that have evolved in response to factors such as the physical constraints of speech production and the cognitive limitations of memory. Furthermore, most formal linguistic frameworks are influenced by literacy and based in  written language; thus, the postulated structures are those that are reflected in the patterns of text.

It is, therefore, unsurprising that many of these linguistic elements are effectively encoded within pre-trained language models trained primarily on next-token prediction or similar learning objectives. Probing classifiers -- simple classifiers designed to map the internal representations of pretrained language models to labels associated with linguistic information -- have demonstrated the ability to extract a wide range of linguistic information, including parts of speech and even parse trees.

\subsubsection{Functional Linguistic Capabilities}
Functional linguistic capabilities, on the other hand, include skills such as logical reasoning, mathematical reasoning, social understanding, and planning. Notably, these capabilities are largely independent of the specific language a speaker uses. This is not to say that there is no variation -- cultural factors play a significant role in shaping how individuals approach these higher-level processes. However, unlike formal linguistic competence, which necessarily varies across languages (with some exceptions), functional linguistic capabilities can remain consistent. For instance, a  bilingual individual can apply the same functional linguistic skills when interacting in different languages while adapting to the formal linguistic rules unique to each language. In this sense, functional linguistic capabilities are, to some extent, language-independent.

Unlike formal linguistic competence, it is not immediately obvious that an LLM trained on next-word prediction would ever develop functional linguistic capabilities. However, effective next-word prediction often necessitates capturing some degree of functional `competence'. For example, consider the sentence: ``The \rule{1cm}{0.15mm} [boring/serene] landscape took our breath away.'' Correctly predicting the word that fits requires some `understanding' -- or at least a statistical approximation -- of what `taking one's breath away' implies. This suggests that LLMs acquire a limited degree of functional linguistic competence through exposure to vast amounts of text.

CxG offers a clear way of distinguishing the functional linguistic competence LLMs might be expected to have or lack.  Under the usage-based acquisition of language, frequency plays a role in a speaker's familiarity with a particular Cxn, as well as the entrenchment of certain instantiations of (i.e. words or morphemes realized within) that Cxn, which in turn plays a role on the extension and generalization of that Cxn. For many literate speakers, there are likely a variety of Cxns that have been and continue to be encountered in text. In these cases, the contextual information that is stored with each exemplar experienced may also be largely textual (although we do not discount the cross-modal associations that may arise, for example, when reading in a comfortable chair, accompanied by coffee and dark chocolate). To the extent that a literate speaker's constructicon is made up of such textual Cxns, it may greatly align with an LLM's constructicon, along with the contextual associations with that Cxn that characterize the stored exemplar. However, where the linguistic information of a speaker will greatly differ from an LLM is in the pre-literate and non-textual encounters with Cxns -- language in real-world communicative interactions. As each token of linguistic experience enriches the stored exemplar of a Cxn, an embodied speaker's constructicon will be characterized by details that are, of course, inaccessible to an LLM. Consider, for example, Labov's department store study: 
speakers demonstrated that they stored socio-economic contextual information with the rhoticized lexical Cxn of, for example `skirt' \citep{labov1986social}. Thus, under a CxG view, LLMs are able to reflect functional linguistic competence insofar as this aligns with a textual context, but are limited in functional linguistic competence that stems from a life-long history of language encounters in an embodied, social context.

Therefore, while LLMs excel at formal linguistic capabilities, their robustness in functional linguistic abilities remains limited, even at the largest scales and when trained on extensive datasets like the entire web. Indeed, several works have highlighted this shortcoming, e.g. \citet{bender2020climbing}. \emph{These limitations highlight the possibility that something is missing, making LLMs particularly ill-suited for the task of `understanding', an important element of functional linguistic competence.}

\subsubsection{High-level, Non-statistically likely Functional Linguistic Capabilities}
Revisiting the example of functional linguistic abilities, it is evident that an LLM would have little difficulty correctly completing the sentence  \textit{``The \rule{1cm}{0.15mm} [boring/serene] landscape took our breath away.''} This demonstrates the need for a finer grained classification of functional linguistic capabilities. Therefore, we propose a finer grained distinction between those elements which typically align with statistical generalizations from text and those that do not. 

This distinction is crucial because the pretraining data of LLMs far exceeds their parametric memory. As a result, LLMs can explicitly store only the most frequently occurring patterns, relying on statistical generalizations to handle less frequent information. This type of generalization -- drawing inferences from priors in the pretraining data without direct `memorization' -- is central to how LLMs address certain functional linguistic tasks. However, the limitations of this approach become apparent when LLMs are tasked with counterfactual reasoning, which requires applying the same underlying rules to scenarios that deviate from standard assumptions. For example, in a counterfactual addition task, the model might need to add numbers in base-9 instead of base-10. 

For this reason, we should expect LLMs to struggle in scenarios where the statistical generalizations from pretraining data do not align with the requirements of the task. Conversely, in cases where these generalizations are the norm -- such as in 
word sense disambiguation -- they can be highly effective. It is for this reason that our evaluation of MWEs in subsequent sections is divided into cases requiring formal linguistic competence and basic, statistically typical functional linguistic abilities, versus those requiring more complex generalizations that cannot be easily derived from pretraining data.

\subsubsection{Meta-linguistic vs Tacit-linguistic knowledge }
\label{sec:meta-tacit}
At this point, it is useful to introduce the concepts of \emph{meta-linguistic information} and \emph{tacit linguistic knowledge}. The latter refers to the intuitive, unspoken understanding native speakers have of their language's Cxns, 
even if they cannot consciously articulate them. This classification is orthogonal to formal and functional linguistic competence as outlined above, and can apply to either. However, traditionally, meta-linguistic information has been more extensively applied in the context of formal linguistic competence. Importantly, testing non-experts on meta-linguistic knowledge -- such as their ability to identify explicit linguistic Cxns -- evaluates only their understanding of formalized concepts, not their tacit knowledge. Assessing tacit linguistic knowledge requires methods that directly examine how these structures are used in natural contexts.

This same distinction applies when evaluating LLMs. Testing whether an LLM can annotate meta-linguistic information primarily assesses its explicit linguistic knowledge rather than its tacit knowledge of language use. It might be argued that, outside of linguistic use cases, what matters most is tacit linguistic knowledge. Therefore, to properly evaluate LLMs, they must be tested in scenarios that reflect real-world language usage.

\subsection{Evaluating the Extent of MWE Generalization in LLMs}

Given this context, we design controlled tests to evaluate the formal and functional tacit-linguistic knowledge of LLMs, specifically focusing on their `understanding' of multiword expressions. These experiments are designed to allow us to assess these models in areas where they are known to perform well and in areas where they are weaker, without relying on meta-linguistic information, which is unlikely to be relevant in real-world scenarios.

Additionally, it is critical to account for the extensive data used during model pretraining. Any meaningful evaluation of a model's high-level abstraction or `understanding' of an MWE must carefully consider the potential influence of pretraining data. Testing whether a model can interpret novel extensions of an MWE in a fashion similar to a human speaker is futile unless we can confirm that those MWEs -- and their variations -- are absent from the pretraining data.

To address this, we conduct tests using novel MWEs that we create, which are therefore, by definition, absent from the pretraining data of LLMs. This ensures that we are evaluating the model's true generalization abilities, independent of any confounding factors introduced by pretraining data exposure.

\subsubsection{Evaluating LLMs' Ability to Track the Linguistic Flexibility of Novel MWEs}
\label{sec:exp1}

\begin{table}[htp]
\footnotesize
\renewcommand{\arraystretch}{1.5}
\begin{center}
\begin{tabular}{ | m{3cm} | m{1.5cm} | m{3cm} | m{3cm} | }
\hline
Context & Statement & GPT\-4o & Annotators   \\
\hline
Expression: to send one’s LLM (instead) \par
Meaning: Not worth one’s time. \par
Usage: There’s no way I’m attending that meeting, I’ll probably send my LLM & 
The bank teller seems to have sent their LLM to work today.  & 
The statement means that the bank teller seems disengaged or inattentive, implying they are not putting in effort or paying attention, as if they had metaphorically `sent their LLM' to do their job instead of being present themselves. & 
Annotator 1: The bank teller is not putting effort into their work, relying on rote memorization of their tasks, thereby making many mistakes and potentially misunderstanding some of their interactions. \\
Expression: \textbf{dance with the deer} \par
Meaning: to have a deep and meaningful conversation, especially about something one is passionate about. \par 
Usage: When they were planning their research, they often danced with the deer. & 
I don’t think they were dancing with the same herd of deer, possibly not even the same species. & 
The statement implies that the individuals involved were having deep and meaningful conversations, but their passions, ideas, or perspectives were significantly different—perhaps even fundamentally incompatible. It suggests a lack of alignment in their discussions or goals. & 
Annotator 1: Although all parties are interested and passionate about their topic of conversation, they have not reached common ground as to what the underlying meaning and implications of what the others are saying.  \\
Expression: Booking a flight for Odysseus \par 
Meaning:  to propose an overly technical or efficient alternative that disregards the deeper purpose, context, or value of the process itself.  \par 
Usage: When Sarah suggested automating the mentoring process with an app, I told her she was booking a flight for Odysseus. & 
The teacher criticized the group’s plan to summarize the novel using AI, saying they were trying to teleport Odysseus. & 
The statement means that the teacher criticized the group for proposing an overly technical or efficient solution (using AI to summarize the novel) that bypasses the deeper purpose or value of engaging with the novel, such as understanding its themes, nuances, and literary depth. & 
Annotator 1: The teacher is conveying that the group's plan is beyond too technical and complicated, and  likely will not work as a result. \\
\hline
\end{tabular}
\end{center}
\caption{\label{tab:eval1} Sample results of our evaluation on novel MWEs that we create. The generalization capabilities of LLMs on this task is noteworthy, given that they could not have been exposed to these MWEs during pretraining. Overall, LLMs are able to correctly infer the meaning of novel uses of all 10 novel MWEs we create. We include one of the two manual annotations here. See the Supplementary Material for full results on all 10 MWEs, including the second annotation.}
\end{table}

\noindent Our first set of experiments are designed to evaluate the extent to which LLMs can identify and generalize novel MWEs defined \emph{in-context} -- within the prompt of the LLM. Notice that these tests are in line with our understanding that LLMs are good at formal linguistic abilities and basic, statistically typical functional linguistic abilities. Therefore, we expect LLMs to be able to be relatively good at this task. As an example, we define the novel MWE `to wink at Pringles': 
\begin{quote}
    \emph{Expression:} To wink at Pringles \\
    \emph{Meaning:} To indulge briefly in something fun or a bit frivolous, almost as a break from more serious things. \\
    \emph{Usage:} They wink at Pringles on Fridays, letting everyone leave early for a quick social hour.''
\end{quote}

\noindent Given the above definition and usage (within the prompt, in the format provided above), an LLM is tasked with interpreting a sentence, which uses this MWE in a different syntactic context, for example: ``I’m so stressed, I need to wink at a lot more than Pringles.'' Notice that this task cannot be solved from memory as the MWE `to wink at Pringles' is a novel MWE that we have created. However, given the definition, we expect the model to be able to generalize the use of such an MWE in novel contexts based on priors extracted from the use of other MWEs in related contexts. 

This exploratory study uses a small dataset consisting of just 10 novel MWEs, created by one of the authors and annotated by the others, who are native speakers of English. We evaluate these MWEs on a single LLM, \textsc{GPT-4o}, one of the latest models available. While the dataset is limited in size, this allows for detailed manual evaluation, and the observed trends are consistent across the examples tested. This, and its extension presented in the next section are not intended as rigorous studies to conclusively demonstrate the exact capabilities of LLMs. Instead, they are intended more as rough guidelines providing evidence for where LLMs break down, and therefore opportunities that we might have in bridging this gap. 

The results, a representative sample of which are presented in Table \ref{tab:eval1} with the complete results included in the supplementary material, reveal a clear pattern: LLMs can generalize the meaning provided in context to novel constructional uses of MWEs. This supports the theory that language learners -- both humans and models -- form generalizations based on exposure to exemplars, a concept central to CxG. Note also that in this study, the human and LLM exposure to the target MWE is the same, given that these are invented MWEs presented for the first time within the instructions for the task. 
These findings 
reinforce the parallels between the importance of frequency of use within a given context in both computational and human linguistic capabilities.  

\subsubsection{The Limits of LLMs: Reasoning with Novel MWEs}
\label{sec:exp2}
Recall our earlier discussion on the varying levels of functional linguistic abilities and how LLMs capture statistically likely functional patterns. The novel MWEs we create offer an ideal opportunity to evaluate LLMs in scenarios that are statistically improbable. Since these MWEs are entirely new, we design a diagnostic test requiring LLMs to combine and reason over information from different novel MWEs. 
This secondary test assesses whether LLMs can generalize independently defined MWE meaning to this level of reasoning over a usage that combines MWEs. For instance, consider the following prompt, which combines two distinct novel MWEs:

\begin{quote}
Using the information given below, output a single character, either A, B, or C, corresponding to the correct answer. Provide an explanation for the answer. If the answer is not clear from the context provided, output the character C.

Consider these novel multiword expression:

Expression: Drown the cables \\
Meaning: Metaphorically severing or overwhelming communication, intentionally or otherwise, so that connection or clarity is lost. \\
Usage: The team drowned the cables by bombarding everyone with conflicting updates.

Expression: dance with the deer \\
Meaning: to have a deep and meaningful conversation, especially about something one is passionate about. \\
Usage: When they were planning their research, they often danced with the deer. 

Now answer this question:\\
Clear instructions are the only requirement for being able to dance with the deer. If the cables were dry, are they likely dancing with different species of deer? \\
a) Yes \\
b) No \\
c) not enough information \\
\end{quote}

We create 10 such reasoning questions, each involving more than one novel MWE. In this experiment, the reasoning questions are presented to two authors and one student volunteer, all native speakers of English. The prompts are presented to the LLM in a `new chat' session, which ensures the model has no prior information pertaining to these MWEs. To improve the chances of the LLM getting the correct answer, the model is allowed to generate a rationale, as prior research has shown that producing a rationale or chain of thought improves accuracy. For this test, we evaluate GPT-o1 and GPT-4o, two of the most advanced LLMs available, trained on a web-scale corpus and containing significantly more parameters than its predecessor, GPT-3, which had 175 billion. We include GPT-o1 since this is a test of `reasoning', and GPT-o1 is known to generate multiple rationales and compare them with each other before generating the final answer. Notably, our aim is not to elicit a single correct answer. Instead, we consider any response correct if it includes a reasonable and non-contradictory rationale. All evaluation is manual to ensure fair and accurate measure of LLM performance. Responses from GPT-4o, GPT-o1 and three human annotators to the above prompt is presented below:

\begin{quote}
    \emph{GPT-4o Response:} \\
    C  \\
    Explanation: The question involves interpreting metaphorical expressions ``dance with the deer'' and ``drown the cables.'' However, the relationship between ``clear instructions'' and ``dancing with the deer'' is unclear. Similarly, the state of ``dry cables'' (absence of communication disruption) does not directly clarify whether they are ``dancing with different species of deer.'' Since the provided information does not establish a clear link between these elements, the correct answer is **C** (not enough information). \\
    \emph{Manually Judged to be incorrect.} \\
    \\
    \emph{GPT-o1 Response:} \\
    Thought about metaphorical expressions for 51 seconds \\
    Answer: B  \\
    Explanation:  From the context, ``dance with the deer'' is a single metaphor meaning to have a deep, meaningful conversation -- and the only stated requirement for that metaphorical ``dance'' is having clear instructions. If the ``cables were dry,'' it simply implies that communication was not overwhelmed or severed (i.e., clarity remained intact). There’s nothing in the provided expressions or their definitions suggesting that a different species of ``deer'' would be involved. Therefore, no, they are not likely dancing with a different species of deer based on the information given.\\
    \emph{Manually Judged to be incorrect.} \\ 
    \\
    \emph{Annotator 1 response:}\\
    C \\
    I'm not convinced of the answer to this one. The statement to me means: Clear instructions are the only requirement for being able to have a meaningful conversation.  If there is no intentional overwhelming of one's communication to be had, they are likely not having meaningful conversations that one would be passionate about. This could be true or not. The question is a bit confusing.
    \emph{Manually Judged to be correct.} \\
    \\
    \emph{Annotator 2 response:}\\
    B\\
    If the cables were dry, meaning communication not broken down, then they should be able to communicate deeply about the subject, given that it is not particularly complex, i.e. requires only clear instructions. 
    \emph{Manually Judged to be correct.}\\
    \\
    \emph{Annotator 3 response:}\\
    C\\
    If the cables are dry, then communication must still be flowing. I think that dancing with a different species of deer means that a deep conversation isn't being had effectively, so if communication is going well, then I wouldn't say that this is likely. \\
    I'm a bit uncertain about the deer metaphor though-- if I am a deer and the deer I'm dancing with are a different species from me, then I get 'different species' as negation. But If I'm not a deer, then I'm not sure I get this negation. Maybe it's enough to be dancing with any deer, and the 'different species' bit should be interpreted in some other way. This feels like a different kind of 'not enough information'-- it's not that the statement isn't clear or the proposition in it isn't supported, it's that I need more info about the metaphor.  
    \emph{Manually Judged to be correct.} \\
\end{quote}

We evaluate each LLM and annotator output individually and mark a response correct if we determine that the justification provided sufficiently backs the response. In this diagnostic test of true generalization, \textbf{GPT-4o correctly respond to 7 out of the 10 questions, and, surprisingly GPT-o1 responds correctly to only 6 of the same questions. Additionally, we find that in these cases, unlike in our previous tests, GPT-4o is particularly brittle and responses vary by quite a bit when presented with slightly different prompts. It's also worth noting that both models were able to correctly infer the meaning of the novel uses of these MWEs in our previous experiment}. By comparison, human annotators, provided with the same prompts, are much more consistent in their responses. While there is some disagreement among the annotators, their justifications remain consistent and in line with their understanding of the novel MWEs. 
Although we emphasize again that our small dataset is suited to a more qualitative analysis in lieu of an empirical evaluation, we also note the following trends in agreement measures: GPT-4o (the best-performing model) matches the ``gold standard'' majority response 58\% of the time, while human annotators agree with each other between 67\% and 100\% of the time. On average, people agree with the gold standard 80.6\% of the time. This shows that people are not only able to provide clear, logical, and importantly, self-consistent reasoning for their responses, but also demonstrate consistency between each other in understanding and reasoning about novel expressions. It's worth noting that two of the three annotators found this specific question challenging, yet their responses still demonstrated a clear understanding of the situation, and their response show this. With the model, however, the issue is different. For example, GPT-o1, after processing the question for 51 seconds, responds with: ``There’s nothing in the provided expressions or their definitions suggesting that a different species of `deer' would be involved.'' This response clearly illustrates that the model fails to make the necessary connection.

Overall, GPT-4o and the arguably more advanced GPT-o1, which sometimes takes close to minute to respond, do NOT demonstrate reasoning that consistently extends to a higher level generalization. The particular lack of robustness to variations in prompts that is more pronounced in these tests highlights this difficulty. While this result is on a particularly small dataset and therefore we cannot say for certain that it is universally generalizable -- across annotators, prompts, or LLMs -- it provides an indication of what LLMs can do and where they begin to fail. It highlights a gap in LLM capabilities when tasked with reasoning beyond statistical priors derived from pretraining data. When reasoning over the combination of novel MWEs, people are able to draw upon the immediate definition of the individual MWE, but also compare the novel Cxns against stored exemplars for similarity, arriving at a richer set of associations to draw from when reasoning over the novel combination.  

\subsubsection{Assessing LLMs on Nascent MWEs}
In addition to our invented MWEs, we conduct an exploratory assessment of three nascent American English MWEs: \textit{got rizz, skibidi toilet,} and \textit{brat summer}. Each of these MWEs is well-attested online over the past two to three years, but is unlikely to be frequently represented in the pretraining data of LLMs, including GPT-4o (although we cannot be sure, as the pre-training data is not disclosed). Thus, we add the following two questions to the secondary reasoning test of both our human annotators and GPT-4o: 

\begin{quote}
 Using the information given below, output a single character, either A, B, or C, corresponding to the correct answer. Provide an explanation for the answer. If the answer is not clear from the context provided, output the character C.

Consider these novel multiword expressions:\\
Expression: Skibidi Toilet\\
Meaning: ``Skibidi'' originally comes from a trending Arabic song. However, it started gaining more recognition from the viral ``Skibidi Toilet'' Youtube Shorts series by Alexey Gerasimov with a storyline around human-headed toilets, according to Forbes. Although ``skibidi'' has no set meaning and can be used interchangeably, the Skibidi Toilet plot portrays it as something evil, giving the word a negative connotation.\\
Usage: He’s so skibidi with that shirt. (X)\\

Expression: Got Rizz\\
Meaning: ``Rizz'' is a term largely used by Gen Z and Gen Alpha and is a shortened version of the word “charisma.''\\
Usage: Just seeing if I still got that W rizz. (TikTok)\\

\textbf{Question 1}: Now, answer this question:\\
Someone yelled, “Bro got that skibidi rizz!” about our teacher. Is it more likely that he should feel…\\
a) Complimented\\
b) Insulted\\
c) Not enough information \\

Consider these novel multiword expressions:\\
Expression: Brat summer\\
Meaning: Dubbed the ``Brat summer'' trend, the confrontational style of Charli XCX’s “Brat” album cover and the specific shade of green became a viral sensation, sparking the brat summer movement that encourages being bold, taking risks, and embracing general “cool girl” style. \\
Usage: Moving from brat summer to demure or mindful fall aligns with most people's internal calendar and their own feelings. (shift.com 17 September 2024) \\

Expression: Got Rizz\\
Meaning: ``Rizz'' is a term largely used by Gen Z and Gen Alpha and is a shortened version of the word “charisma.''\\
Usage: Just seeing if I still got that W rizz. (TikTok)\\

\textbf{Question 2}: Now, answer this question:\\
If Charli XCX’s last album was brat summer, the latest remix album is bratumn because of its distinct Autumn rizz and not just because it was released in October. The remix album is likely…\\
a) Darker and more detached\\
b) Even more confrontational\\
c) Not enough information \\
\end{quote}

\noindent For both of the above questions, GPT-4o does not provide the gold-standard, majority response from human annotators (C: there is not enough information to determine if the teacher should be complimented or insulted, and A: the remix album is likely darker and more detached), nor does it provide a cogent explanation for its reasoning. 

These actual, nascent MWEs reflect the complexity of real MWEs, where the meaning must be understood with respect to the cross-modal contextual information that comes along with encountered exemplars. In the case of \textit{Skibidi Toilet}, the meaning is informed by the relatively jarring visuals of the toilet with a head emerging from it that accompanied the videos that introduced the MWE to the lexicon. Perhaps even more interestingly, the word \textit{Skibidi} itself is an example of onomatopoeia, capturing the vocalization of a beat to a song that also accompanied the video.  In the case of \textit{Brat Summer}, the meaning is informed by a variety of cross-modal characteristics of Charli XCX's album, ``Brat,'' including the bright green album cover as well as the confrontational but can't-be-bothered feel of the music. Thus, for people with these Cxns, the stored exemplars are extremely rich with a variety of multi-sensory features. Speakers tap into these stored, cross-modal details when reasoning over the question.  This is clearly distinct from any LLM representation of these Cxns, which are limited to textual context only. 

\subsubsection{Implications to Future Research in MWEs and NLP}

Notice that our first set of experiments (Section \ref{sec:exp1}) observed that LLMs \emph{can} answer fairly complex questions pertaining to novel MWEs correctly, even when these involve novel use of previously unseen MWEs. However, their performance degradation in this new setting, when tested on the very same MWEs, suggests a reliance on single-step inference grounded in pretraining priors, rather than a robust capacity for high-level generalization and reasoning.

This distinction is important. The failure to answer the second set of questions, and particularly, the small, but surprising drop in performance of GPT-o1 combined with the lack of robustness to prompt variations of GPT-4o, indicates that LLMs are using a combination of statistical priors from pretraining and information from the prompt to approximate plausible responses. These priors enable them to perform well in scenarios where the task is in line with familiar patterns (or analogies) from their training data. However, when faced with tasks requiring a higher level of generalization, their limitations become apparent.

From a CxG perspective, given the importance of frequency of use, arguably you as a natural language speaker are also using statistical priors from your own experience with language in order to `understand' language. Where this experience is similar for a given Cxn, as in the first experiment, the linguistic abilities of people and LLMs is also surprisingly similar. That is, when the level of generalization requires only looking back to one previous token of textual use, people and LLMs can perform comparably. However, the reasoning questions in the second experiment push for generalization of that single attestation to very different contexts, such as counterfactuals (for example, generalizing from the meaning of ``drown the cables'' to ``if the cables were dry...''). Here, speakers tap into a lifetime of experience with embodied concepts such as wet/dry, the effect of wetness on electrical cables, as well as cross-modal contextual and cultural associations with semantically similar Cxns as metaphors, such as ``losing the thread'' of a particular topic.  These detailed associations allow speakers to consistently reason over the question, while LLMs lack comparable, real-world associations.

While it is notable that LLMs successfully solve 70\% of our relatively small dataset, it remains unclear how to consistently improve these models, given their opacity and immense scale. Notice that the more powerful GPT-o1 does \emph{worse} than GPT-4o, and also does not get the same questions right. It is at this juncture that we believe that MWE research has a unique opportunity to contribute. By integrating insights from CxG, MWE research can offer interpretable and explainable datasets and methods. One such method might be the incorporation of AMR or UMR annotation schema into the MWE annotations. These resources can build upon the impressive formal and low-level functional capabilities of LLMs as a foundation, scaffolding toward systems that are more robust and generalizable. We believe this approach represents a more sustainable path forward.

\section{Future Directions \& Conclusions}
\label{sec:conclusion}
This research demonstrates the standing value of having linguistic resources like PropBank and UMR for modeling the properties of MWEs. With such resources, we can better understand the patterns of flexibility and productivity. This also facilitates recognition and interpretation of MWEs in low-resource languages, where LLMs simply may not ever be feasible. However, such resources are also important in English, as they  provide evaluation data with some explanatory power as to what linguistic capabilities LLMs seem to have and which they are lacking. As our experiments have shown here, although an LLM's ``knowledge'' of language does seem to have surprising parallels to usage-based acquisition of a constructicon, LLMs remain unable to generalize Cxns in a fashion similar to people, potentially because our extension of Cxns generalizes over a much richer set of cross-modal features than that of a text-based LLM. 

We have shown that CxG provides a unified way of thinking about MWEs cross-linguistically even when the languages differ greatly with respect to the notion of wordhood.  Furthermore, Cxns offer a powerful template for modeling the flexibility of MWEs as well as providing a better understanding their partial productivity. Developing constructional resources, such as PropBank and UMR, that mark up the fixed and flexible slots of constructions can also facilitate both theoretical and computational analysis of the range of words that fill certain slots.  

To the extent that CxG aligns with LLMs as a usage-based theory of language acquisition and grammar, this elucidates why LLMs do seem to have a powerful grasp of language. On the other hand, where CxG highlights the cross-modal nature of the information stored for each token of linguistic experience, this also demonstrates the expected demarcation between an LLM's knowledge of language in contrast to a person's. With this awareness, we can begin to bridge the gap between our linguistic abilities and that our NLP systems' by pushing forward research into how to encode cross-modal features. While part of the answer may come in the form of larger, multi-modal pretraining corpora, we also emphasize the power of symbolic resources in providing explicit form-meaning constructional associations, in order to scaffold the bridge to future functional linguistic competence in our language technologies. Above all, we join others such as \citet{bisk2020experience} in arguing that in order to process language and arrive at the same interpretation as a person, NLP resources must widen scope from a focus on text and even images, to aspects of embodied social interaction.

{\sloppy\printbibliography[heading=subbibliography,notkeyword=this]}

\end{document}